\documentclass[letterpaper, 10 pt, conference]{ieeeconf}  

\IEEEoverridecommandlockouts                              
\usepackage{amsmath}
\usepackage{graphicx}
\usepackage[caption=false]{subfig}
\usepackage{multirow}
\usepackage{dblfloatfix}
\usepackage{siunitx}
\usepackage{booktabs}
\usepackage{longtable}

\usepackage{authblk}

\usepackage{nomencl}
\makenomenclature

\title{\LARGE \bf
CoHaptics: Development of Human-Robot Collaborative System with Forearm-worn Haptic Display to Increase Safety in Future Factories

}

\author{
Miguel Altamirano Cabrera$^{1}$,
Juan Heredia$^{2}$,
Jonathan Tirado$^{1}$, \\ 
Vladislav Panov$^{3}$,
Fikre Hagos$^{1}$,
Dzmitry Tsetserukou$^{1}$
\thanks{$^{1}$ The authors are with the Space Center, Skolkovo Institute of Science and Technology (Skoltech), 121205 Bolshoy Boulevard 30, bld. 1, Moscow, Russia. {\tt\small \{miguel.altamirano, jonathan.tirado, fikre.hagos, d.tsetserukou\}@skoltech.ru\ }}

\thanks{$^{2}$ The author is with The Maersk Mc-Kinney Moller Institute at the University of Southern Denmark, Odense, Denmark. {\tt\small jehm@mmmi.sdu.dk}}

\thanks{$^{3}$ The author is with PMAR Robotics Group, Department of Mechanical, Energy, Management, and Transportation Engineering, University of Genoa, 16145 Genoa (GE), Italy. {\tt\small S4957274@studenti.unige.it}}%
}

\begin{document}
\maketitle
\thispagestyle{empty}
\pagestyle{empty}

\begin{abstract}

Complex tasks require human collaboration since robots do not have enough dexterity. However, robots are still used as instruments and not as collaborative systems. We are introducing a framework to ensure safety in a human-robot collaborative environment. The system is composed of a haptic feedback display, low-cost wearable mocap, and a new collision avoidance algorithm based on the Artificial Potential Fields (APF). Wearable optical motion capturing system enables tracking the human hand position with high accuracy and low latency on large working areas. This study evaluates whether haptic feedback improves safety in human-robot collaboration. Three experiments were carried out to evaluate the performance of the proposed system. The first one evaluated human responses to the haptic device during interaction with the Robot Tool Center Point (TCP). The second experiment analyzed human-robot behavior during an imminent collision. The third experiment evaluated the system in a collaborative activity in a shared working environment. This study had shown that when haptic feedback in the control loop was included, the safe distance (minimum robot-obstacle distance) increased by 4.1 cm from 12.39 cm to 16.55 cm, and the robot's path, when the collision avoidance algorithm was activated, was reduced by 81\%.

\end{abstract}

\section{Introduction}

The implementation of robots in industrial and service sectors has been increasing in the last years. However, in the majority of domains, robots are still implemented more as tools than as systems that can collaborate with humans in the same working area. Close collaboration between robots and humans requires a flexible, fast, and safety-oriented control capable of ensuring human protection in dynamically changing environments \cite{Bicchi2008}.

Traditionally, industrial robots work inside cages due to safety protocols. The guidelines claimed that humans and industrial robots could not coexist in the same environment. However, modern literature on Human-Robot Interaction has highlighted several frameworks where robots and humans coexist. The simple strategy for coexistence is to stop the robot when humans are near the robot's path area \cite{paper1, paper2}. This method, however, does not allow collaboration. Besides, other authors have proposed collaborative systems where the robot can dynamically avoid collisions with humans.
\begin{figure}[ht]
  \centering
  \includegraphics[width=0.46\textwidth]{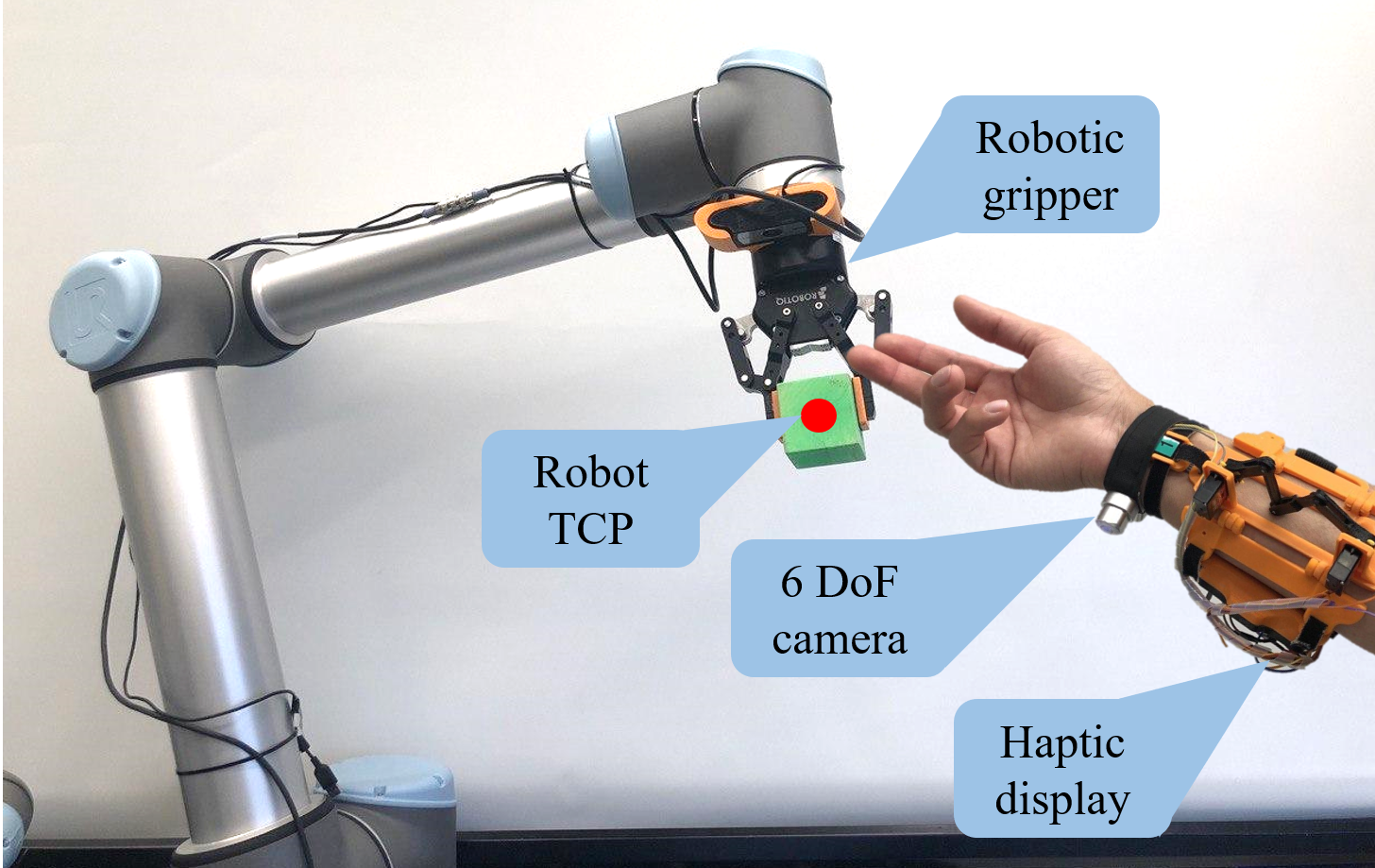}
  \qquad
  \caption{Layout of the collaborative robotic system CoHaptics. User's wrist is tracked by the wearable  Antilatency mocap. Novel wearable haptic device $RecyGlide$ generates tactile patterns on the forearm. }
  \label{fig:SystemOverview}
  \vspace{-3mm}
\end{figure}

This paper presents a novel system framework to guarantee safety during human-robot collaborative tasks. A new framework is composed of a wearable multi-modal haptic display $RecyGlide$, wearable mocap by Antilatency, and an improved collision avoidance algorithm based on the Artificial Potential Fields (APF). The wearable haptic display aims to inform the users about their distance to the robot Tool Center Point (TCP) and  robot collision avoidance algorithm mode. This haptic information will reduce the activation of the avoidance controller, decreasing the operation time, and improving human safety. 

\section{Related Works}
The concept of collision avoidance in robotics is proposed in \cite{paper1, paper2} where the robot avoids obstacles using a camera for environment perception and novel algorithms.  In \cite{paper3}, authors present an algorithm based on neural networks to detect collisions using a force sensor at the robot TCP.  The approaches differ in the perception technology and the algorithms for collision avoidance. 
                     
To allow human-robot collaboration, the robot is equipped with sensors to detect the collision with humans. For example, force-torque sensors have been used at the robot TCP \cite{Siciliano2019}, or at the base \cite{Das}. In \cite{Tsetserukou}, an obstacle avoidance control through tactile perception using optical torque sensors was introduced. However, the interaction with the environment was limited due to the contact of the robotic arm with obstacles.  Because all off-the-shelf collaborative robots are equipped only with force/torque sensors, they stop after the collision is detected. Therefore, to guarantee a high level of safety, we must generate the collision-free trajectories in advance to avoid any contact with the human.

Environment perception is used to detect the positions of the obstacles in real time using different methods. The implementation Computer Vision (CV) algorithms for obstacle avoidance has been done in \cite{Cuevas-Velasquez2018}, providing a system for hybrid visual servoing to moving targets. However, the developed system requires careful calibration, and potentially, the data acquisition can be affected by latency due to the long processing time of each image frame. 

Nowadays, passive optical mocap, such as VICON, OptiTrack, or NDI's Polaris, is the most popular solution for environment perception. Their advantages are low latency and high accuracy. However, such systems require a powerful computer for data processing, have a limited working area, and are very expensive. Additionally, passive markers can be occluded by the robot moving above the human hand. We propose to apply 6 Degrees of Freedom (DOF) wearable optical mocap (by Antilatency) featuring onboard data processing with Field-Programmable Gate Array (FPGA) resulting in low latency of 2 ms. Importantly, the required working area can be easily arranged by placing infrared markers on the floor. In case the wearable camera is temporally occluded, the hand position would be available using the Inertial Measirement Unit (IMU) odometry.   

The Artificial Potential Field (APF) approach introduced in \cite{Khatib} proposes virtual repulsive and attractive fields associated with obstacles and targets faced during robot movement. This approach was implemented in autonomous mobile robots. It was further elaborated in \cite{SGe}, introducing new repulsive potential functions to avoid the local minimum problem, ensuring that the robot reached the goal. In \cite{HLin}, a repulsive force field in three-dimensional space was presented to adjust the original trajectory of the robot and to avoid obstacles smoothly. The idea of using potential field with a fictional repulsive force around the workspace obstacles was suggested in \cite{Liao} and later developed in \cite{Luecke}.

A number of studies have begun to examine sensory augmentation to improve the user's performance.  Recent research has focused on using sensing technologies to capture the human position and to deliver this information to the human brain through Human Machine Interfaces (HMI). After training, the user may acquire this additional information and process it as an extra sense. For example, in \cite{Park2017}, a framework of automated safety monitoring  was proposed in which the position of the worker was tracked and hazard alerts were sent to prevent possible accidents in an industrial environment. In a similar case, \cite{Cho2018} presented a haptic device for worker safety. They studied pattern recognition and how these patterns modify their behavior in a determined activity.

Wearable haptic interfaces have been used to deliver information to users in dangerous or dexterous conditions to improve the user's performance, such as in teleoperation of robots. The implementation of haptic feedback during teleoperation provides many advantages, such as increasing the spatiotemporal coordination in collaborative environments and the situational awareness of the operator \cite{Ang2015}. In \cite{Bimbo} a robotic teleoperation system with wearable haptic feedback for telemanipulation in cluttered environments was presented, where the collision with object was rendered on the upper arm and forearm by vibrotactile stimuli. Furthermore,  \cite{PacchierottieKuchenbecker} introduced cutaneous feedback of finger deformation and vibration in the teleoperation of robotic surgery to increase the telepresence illusion of the surgeon. Such procedures require high accuracy of haptic information transmission. While working with collaborative robots human vision is overloaded and haptic interfaces providing additional channel can potentially improve users’ safety during work.

\section{System Architecture} \label{System Integration}

\begin{figure}[ht]
  \centering
  \includegraphics[width=0.45\textwidth]{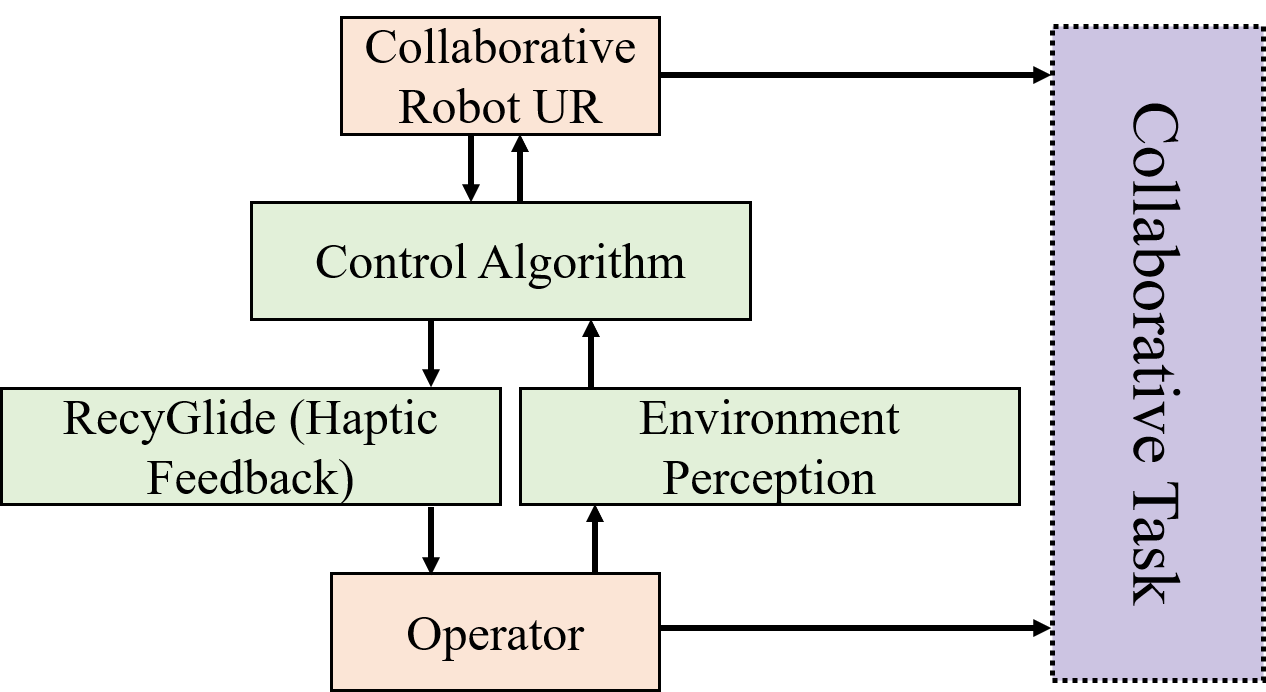}
  \qquad
  \caption{CoHaptics system layout.} 
  \label{fig:SystemIntegration}
  \vspace{-3mm}
\end{figure}

The system is composed of three modules: environment perception, control algorithm, and haptic feedback, as is shown in Fig. \ref{fig:SystemIntegration}. The collaborative robot and operator work on a common collaborative task. The $control$ $algorithm$ guides the robot TCP to the desired position avoiding obstacle (operator). The $environment$ $perception$ system informs the control algorithm on the user position. The previously mentioned structure was presented, tested, and evaluated in \cite{9217041}. In this paper, we include $haptic$ $feedback$ which communicates the position of the robot to the operator. The operator experiences the haptic stimuli on the forearm rendered by the haptic display $RecyGlide$.

\subsection{Environment Perception}

Antilatency technology is a novel wearable visual-inertial mocap system composed of passive beacons and a tracking device. Reference stripes with active infrared markers are located on the floor, and a wireless tracking device is attached to the tracked object. The tracking device calculates the position and orientation using a 6 DOF wearable optical camera and an IMU sensor. While being temporally occluded, the position is calculated by IMU odometry algorithms. We propose to apply this novel technology, which has been used only in virtual reality environments, to the industry. 

The hand position data of the user is sent from the Antilatency device, located in the user's wrist, to our perception algorithm via a wireless radio socket connection each 2 ms. The user freely moves around the working environment and interacts with the robot. The coordinate system of the Antilatency device is transformed to the robot coordinate system using rotational and translational coordinate transformations.

\subsection{Control Algorithm}

The control algorithm was proposed by us in \cite{9217041}. We describe a brief review of the controller algorithm in this subsection. The controller moves the robot TCP position ${x}_{R}$ to the goal position ${x}_{G}$ avoiding the obstacle position ${x}_{O}$ based on the APF approach, which proposes  virtual repulsive and attractive fields associated with obstacles and targets faced during robot movement. The operative frequency of the algorithm equals 100 Hz.

 \begin{figure}[ht]
  \centering
  \includegraphics[width=0.45\textwidth]{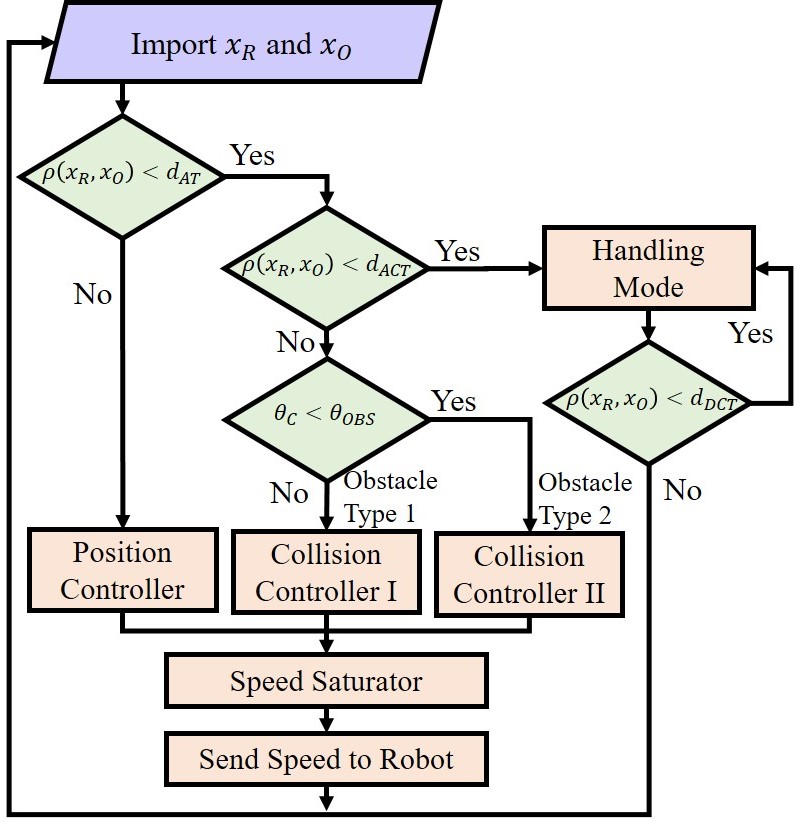}
  \qquad
  \caption{Behavior tree of the control architecture. Position controller and collision controller I and II are defined by (\ref{eq:C}), (\ref{eq:CI}), and (\ref{eq:CII}), respectively. } 
  \label{fig:Three}
\end{figure}

We apply the algorithm represented in the behavior tree Fig. \ref{fig:Three}. The decisions are made based on the distance between the robot TCP and the obstacle $d_{RO} = \rho(\boldsymbol{x}_{R}, \boldsymbol{x}_{O})$. The behavior tree is divided into three cases. 

\subsubsection{Case 1} The obstacle is far from the robot TCP position. The variable $d_{RO}$  is larger than the avoidance threshold distance $d_{AT}$. Therefore, the risk of collision is low. The algorithm employs the position controller represented in equation (\ref{eq:C}):
\begin{equation}\label{eq:C} \boldsymbol{\dot{q}}_{PC} = J^{-1} \cdot ( \boldsymbol{\dot{x}}_{G} + k_{PC1} \tanh{(k_{PC2} \cdot \boldsymbol{e})})   , \end{equation}
where $\dot{q}_{PC}$ is the articulation speed of the TCP, $J$ is the Jacobian matrix, $\boldsymbol{e}$ is the position error between the goal position ${x}_{G}$ and the robot TCP position ${x}_{R}$, $k_{PC1}$ and $k_{PC2}$ are the calibration parameters of the collision avoidance controller. 

\subsubsection{Case 2} The obstacle is inside the avoidance area ($d_{RO} < d_{AT}$), but not in the critical area ($d_{RO} > d_{ACT}$). In such a case, there is the risk of collision. The algorithm classifies the obstacle as one of two different types according to the angle between the TCP velocity vector and the robot-obstacle vector $\theta_{C}$. When the angle $\theta_{C}$ is less than the threshold value $\theta_{OBS}$ the obstacle is type 1, and the collision is imminent if the robot movement continues in the same direction. If the angle $\theta_{C}$ is more than the threshold angle $\theta_{OBS}$, the obstacle is type 2, which means that robot is moving away of the obstacle. Equations (\ref{eq:CI}) and (\ref{eq:CII}) defines the obstacle avoidance control in the case of obstacle type 1 and obstacle type 2, respectively: 
 \begin{equation}
\label{eq:CI}
\dot{q}_{CCI} = J^{-1}( \boldsymbol{v}_{PC} ( 1 - e^{-\tau d_{RO}}) + \boldsymbol{v}_{rep_1}  e^{-\tau d_{RO}})   , 
\end{equation}
where $\tau$ represents the space-null attenuation constant, ${v}_{PC}$ and ${v}_{rep_1}$ are the speed of position controller and speed of the collision controller I, respectively, ${v}_{rep_1}$ is composed of three repulsive forces to change the direction of the robot TCP and avoid the obstacle \cite{9217041}.
\begin{equation}
\label{eq:CII}
\dot{q}_{CCII} = J^{-1}( \boldsymbol{v}_{PC} ( 1 - e^{-\tau d_{RO} }) + \boldsymbol{v}_{rep_2}  e^{-\tau d_{RO}}) ,
\end{equation}
where $\tau$ represents the space-null attenuation constant, ${v}_{rep_2}$ is the speed of the collision controller II that is composed of only a normal repulsive force \cite{9217041}.

\subsubsection{Case 3}  When $d_{RO}$ is smaller than the critical threshold distance of activation $d_{ACT}$ obstacle is exceptionally close to the robot TCP position. Therefore, the risk of collision is high. The algorithm initiates a Free Drive Control Mode (FDCM). In the case of FDCM, the robot stops and the user can modify the robot's position by pulling-pushing the end effector. The system deactivates FDCM when the robot-operator distance $d_{RO}$ is larger than the critical threshold value of deactivation $d_{DCT}$.

\nomenclature{$\theta_{OBS}$}{Obstacle Angle: Threshold angle to distinguish between Case 1 or Case 2}
\nomenclature{$d_{DCT}$}{Critical avoidance distance: Parameter that deactivates the handling mode control}
\nomenclature{$d_{ACT}$}{Critical avoidance distance of activation: Parameter that activates the handling mode control }
\nomenclature{$d_{AT}$}{Avoidance threshold distance: Parameter that activates the avoidance controllers}
\nomenclature{$\theta_{C}$}{Angle between the robot velocity vector and the robot-obstacle vector}
\nomenclature{$d_{RO}$}{Distance between robot and obstacle}
\nomenclature{${x}_{O}$}{Operator's hand position}
\nomenclature{${x}_{G}$}{Desired goal position}
\nomenclature{${x}_{R}$}{Position of the robot TCP}

\subsection{Haptic Display}

\begin{figure}[ht]
  \centering
  \includegraphics[width=0.48\textwidth]{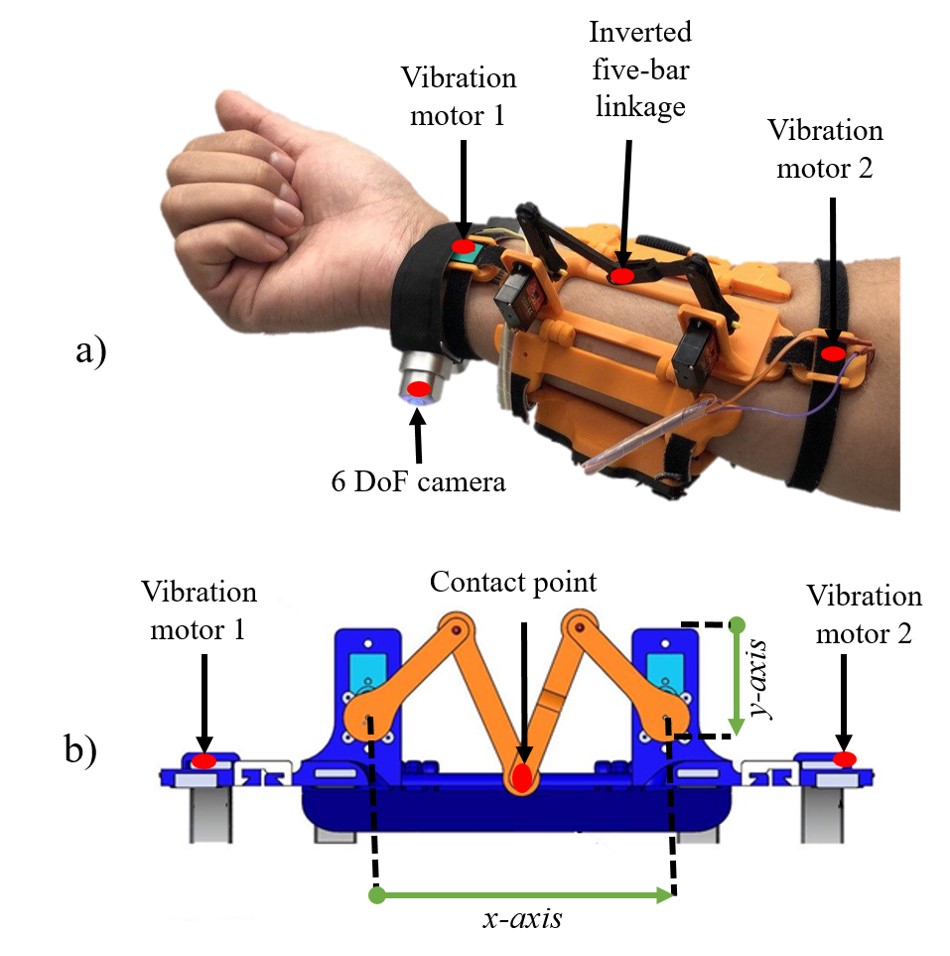}
  \qquad
  \vspace*{-6mm}
  \caption{$RecyGlide$, a novel wearable haptic device with inverted five-bar linkage and vibration motors for delivering multi-modal tactile stimuli $RecyGlide$. a) Prototype of $RecyGlide$ worn on the user forearm. b) CAD model. } 
  \label{fig:RecyGlide}
  \vspace{-3mm}
\end{figure}

$RecyGlide$ is a novel wearable tactile display of 95 $g$ to deliver multi-modal stimuli at the user's forearm \cite{10.1145/3359996.3364759}. The device provides the sense of touch at one contact point by an inverse five-bar linkage mechanism inspired by $LinkTouch$ technology \cite{tsetserukou2014} and two vibration motors. $RecyGlide$ device worn on the human forearm and CAD model are presented in Fig. \ref{fig:RecyGlide}(a) and Fig. \ref{fig:RecyGlide}(b), respectively. The device operative frequency is 100 [Hz]. The vibration cue provides alert commands to the user, while inverse five-bar linkage renders a linear displacement sensation. The combination of these two stimuli generates haptic feedback to the human.  

To achieve the contact point sensation in forearm longitudinal direction, the device has two servo motors which are generating the target joint angles of the inverse five-bar linkage mechanism. The $RecyGlide$ provides a maximum 75 mm displacement of the contact point along the forearm (Fig. \ref{fig:RecyGlide}(b)). A maximum normal force at the contact point is equal to 2 $N$.

The second stimulus is generated by two vibration motors located on the extreme sides of the device. The two types of stimuli allow the designing of different tactile patterns at the forearm. The device has been designed to adapt ergonomically to the user's forearm, allowing free movement of the hand when hands are used for object manipulation.

$RecyGlide$ is composed of an ESP32 microcontroller, two servomotors PowerHD DSM44 (motor torque 0.12 N/m), and two coin vibration motors of 8 mm diameter and 3.4 mm thickness. The ESP32 is responsible for controlling the angles of servo and vibration motors, and for Bluetooth communication to the computer.

\begin{figure}[ht]
  \centering
  \includegraphics[width=0.48\textwidth]{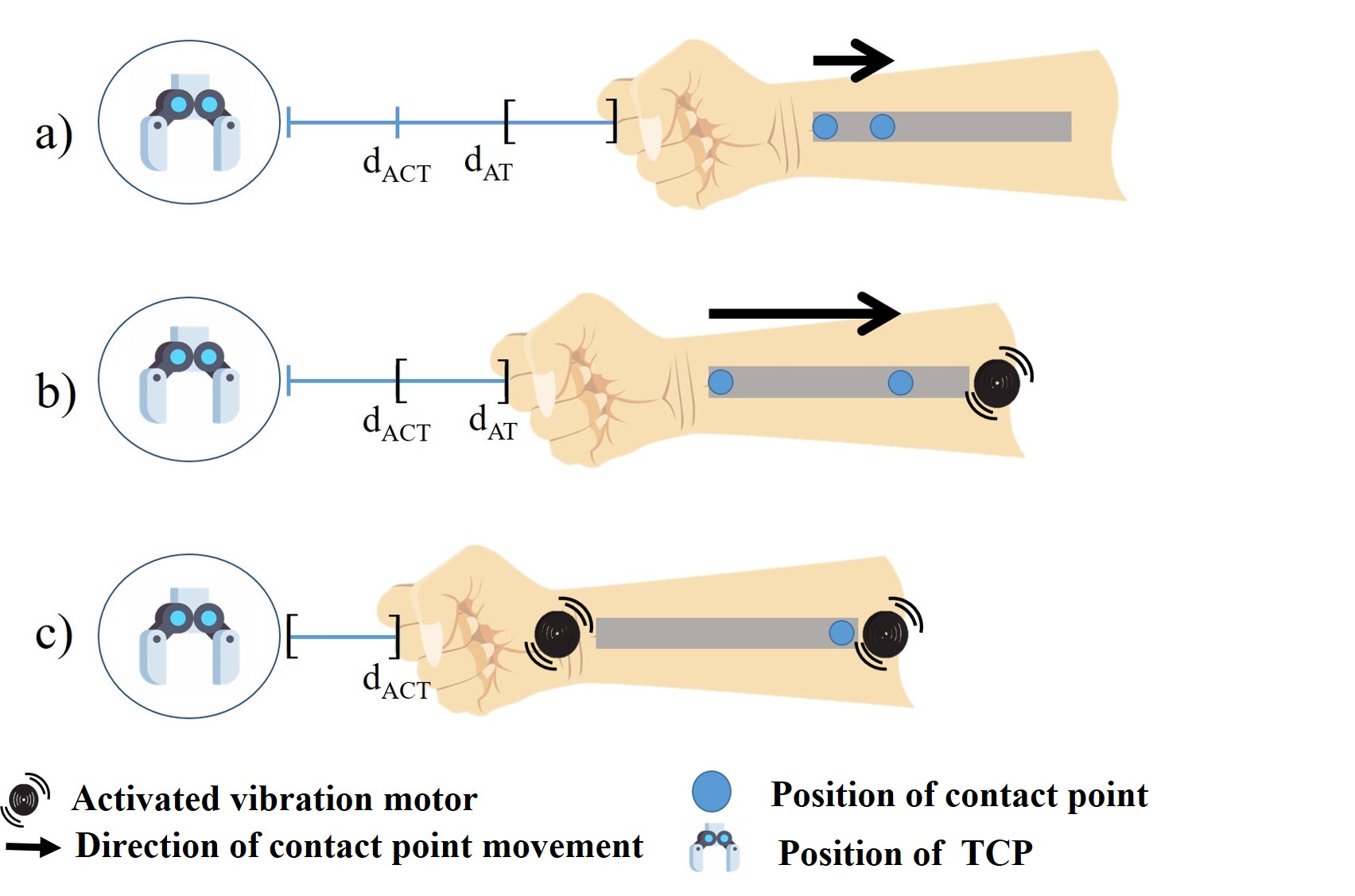}
  \qquad
  \vspace*{-4mm}
  \caption{Multimodal haptic patterns for: a) $d_{RO} > d_{AT}$. b) $d_{ACT} < d_{RO} < d_{AT}$. c) $d_{RO} < d_{ACT}$. } 
  \label{fig:Rendering}
  \vspace{-3mm}
\end{figure}

The novelty of this work is to analyze the influence of haptic feedback provided by the proposed device during a human-robot interaction task. In our hypotheses, bidirectional communication between the operator and the system reduces the security hazard and decreases the robot's path while activating the collision avoidance. The haptic device renders multi-modal stimuli according to three interaction modes:

\begin{itemize}
\item The distance from the robot TCP to the user's hand $d_{RO}$ is larger than the avoidance threshold distance $d_{AT}$, as shown in Fig. \ref{fig:Rendering}(a). The distance between the TCP and the user's hand $d_{RO}$ is rendered by the inverse five-bar mechanism at the operator's forearm.

\item The distance from the TCP to the user's hand $d_{RO}$ is between the critical distance of activation $d_{ACT}$ and the avoidance threshold distance $d_{AT}$, as shown in Fig. \ref{fig:Rendering}(b). In this area, the collision controller I and collision controller II are active. The distance $d_{RO}$ continues to be rendered by the sliding of the contact point, and one vibration motor of the device gets active at 40\% of its maximum power. 

\item The distance from the TCP to the user's hand $d_{RO}$ is smaller than the critical distance of activation $d_{ACT}$ as shown in Fig. \ref{fig:Rendering}(c). FDCM is activated to avoid collision with the user. The two vibration motors of the device are active at 70\% of their maximum power.  
 
\end{itemize}

\section{Experimental Evaluation} \label{ALGORITHM CALIBRATION EXPERIMENTS}

Two experiments were carried out to evaluate the performance of the proposed system. The first experiment evaluates the human responses to the haptic stimuli. The second experiment analyzes human and robotic behavior during an imminent collision. 

\subsection{Experiment 1}

This experiment evaluates the behavior of the user under different tactile stimuli generated by haptic display $RecyGlide$. The experiment determines a spatial relation between the rendering range of the device and the real robot-operator distance $d_{RO}$.

The experiment was executed in a structured environment. The user was asked to wear an Antilatency tracker and $RecyGlide$ on the right forearm. We covered the user's eyes, and asked to go around a static point, which emulates the robot TCP presence, keeping constant distance. The distance between the position of the Antilatency tracker and the static point was rendered to the user's forearm by the displacement of the contact point. 

The rendering range length $d_{SR}$ determines the limit rendering distance. When the distance between the user and the static point $d_{RO}$ is larger than the rendering range length $d_{SR}$, the haptic device does not provide any feedback. When the distance between the user and the static point $d_{RO}$ is smaller than $d_{SR}$, the haptic device renders the distance $d_{RO}$ to the user's forearm by displacement of the device's contact point. For this experiment we proposed four different rendering range lengths, which are 5 cm (Case 1),10 cm (Case 2), 15 cm (Case 3) and 20 cm (Case 4). The proposed rendering range lengths are proportionally scaled to the maximum displacement length of the haptic device contact point, which equals 7.5 cm. 

\begin{figure}[h!]
  \centering
   \includegraphics[width=0.5\textwidth]{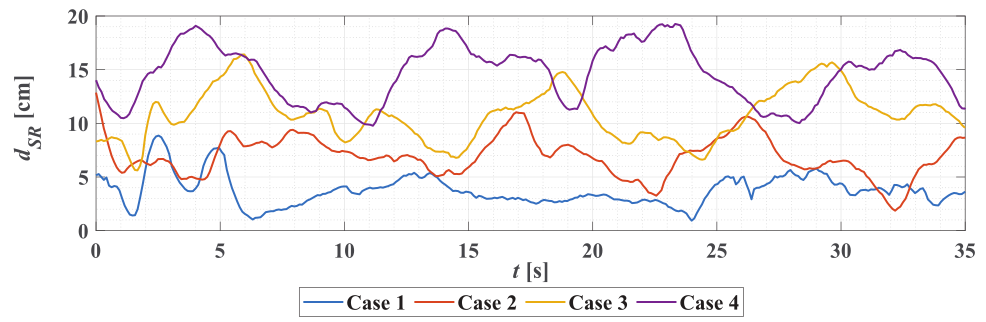}
  \qquad
  \vspace*{-5mm}
  \caption{Distance from the user to static point $d_{A}$ for each of the four rendering range values cases, where the case 1 corresponds to 5 cm, case 2 to 10 cm, case 3 to 15 cm, and case 4 to 20 cm.}
  \label{fig:dmin3}
  \vspace{-3mm}
\end{figure}

\begin{figure}[h!]
  \centering
  \vspace*{0.2cm}
  \includegraphics[width=0.4\textwidth]{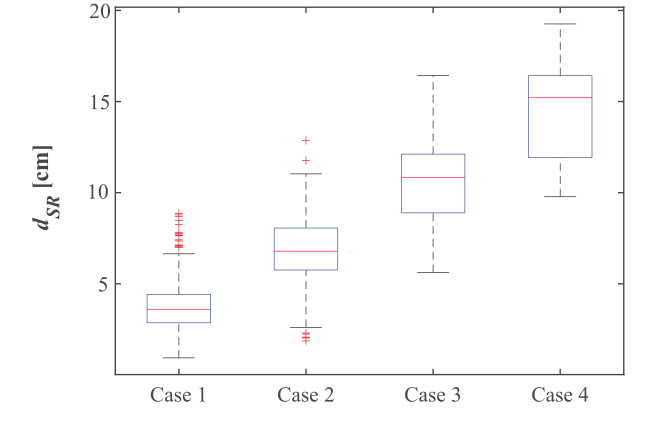}
  \qquad
  \vspace*{-5mm}
  \caption{Distance from the user to static point $d_{A}$ of the four experimental cases: (Case 1: 5 cm) (Case 2: 10 cm) (Case 3: 15 cm) (Case 4: 20 cm).}
  \label{fig:dmin2}
  \vspace{-3mm}
\end{figure}

\textbf{Results}: Fig. \ref{fig:dmin3} shows the distance between the user and static point $d_{A}$ for the four cases versus time. Additionally, Fig. \ref{fig:dmin2} shows the whisker plot of each case. The relation between the medians of each case is median (case1) $<$ median (case2) $<$ median (case3) $<$ median (case4). The interquartile range (IQR) increases from each case to the next one, and the relation is IQR (case1) $<$ IQR (case2) $<$ IQR (case3) $<$ IQR (case4).

Fig. \ref{fig:dmin2} shows that the median and deviation of $d_{RO}$ increase when the range length $d_{SR}$ increases. In case 4, the user is further away from the static point compared to the other 3 cases. We observe that the user did not overpass the threshold of 0.1 m during the case 4.  

The experiment shows that a blind user is capable of avoiding a static point using only haptic feedback. This result shows that the implementation of haptic feedback can potentially reduce the risk of collision between the user and an object, which can be the robot TCP.

\subsection{Experiment 2}

The second experiment aims at analyzing human and robotic behavior during an imminent collision. To observe how the robot behaves with the collision avoidance algorithm, in a first instance, the user's hand was located statically in the middle of the robot TCP trajectory from point A to point B without haptic feedback. In a second instance, the user's hand was located in the same position as in the first instance, and the haptic display was activated. The user was asked to move his hand according to the perceived sensation to avoid the robot TCP.  

The critical distance of activation $d_{ACT}$ was set at 10 cm, and the avoidance threshold distance $d_{AT}$ was set at 30 cm according to the previous calibration in \cite{9217041}. The haptic display was set to render the distance from the robot to the user's hand $d_{RO}$ in a range of 40 cm to 20 cm, as in case 4 of experiment 2, to avoid being closer than the critical distance of activation.

\begin{figure}[h!]
  \centering
  \includegraphics[width=0.45\textwidth]{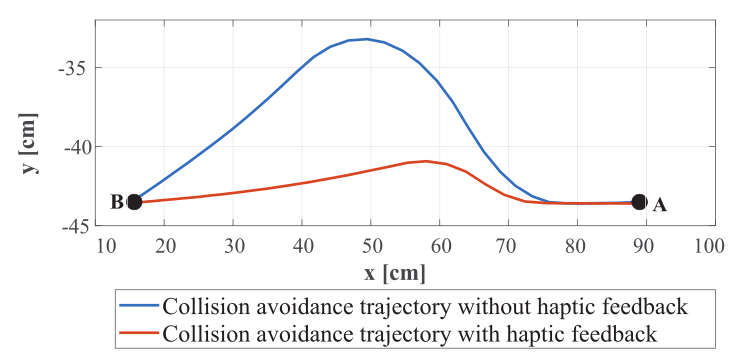}
  \qquad
  \vspace*{-5mm}
  \caption{Experiment 2: the robot TCP path curvature when interacting with the static user's hand without haptic feedback (blue line), and with haptic feedback (red line) to avoid the robot TCP.}
  \label{fig:dmin1}
\end{figure}

\textbf{Results}: Fig. \ref{fig:dmin1} shows the behavior of the system when the user's hand is static without haptic feedback (collision avoidance trajectory, blue line), and with haptic feedback (orange line). From robot point of view, the user with haptic feedback behaves as a dynamic obstacle, modifying the user-robot distance constantly. We can observe that the TCP did not collide with the user's hand when it was static. Furthermore, the trajectory of the robot TCP, when the haptic feedback was active, was less than the path when the user's hand was static. This is because the user started to be informed that the robot was close when the TCP was 40 cm to the user, and he/she had the time to move the hand and avoid a closer distance. 
With these results, we can conclude that the robot's path can be reduced if we provide haptic feedback to the user, increasing the user-robot distance and decreasing the activation of the collision avoidance algorithm. 

$RecyGlide$ enhances the perception of the robot's position and constantly informs the user to avoid the robot TCP, and to be aware of possible collisions. $RecyGlide$ provides information to allow collision avoidance from the user size, reducing the robot's path and execution time. 

\section{Human-Robot Interaction Experiment in a Collaborative Task}

In preliminary experiments, each part of the system was tested. In \cite{9217041}, the controller was calibrated and assessed through three experiments. This experiment aimed to test and evaluate the system structure and integration in a collaborative task. 

\begin{figure}[ht]
  \centering
  \includegraphics[width=0.49\textwidth]{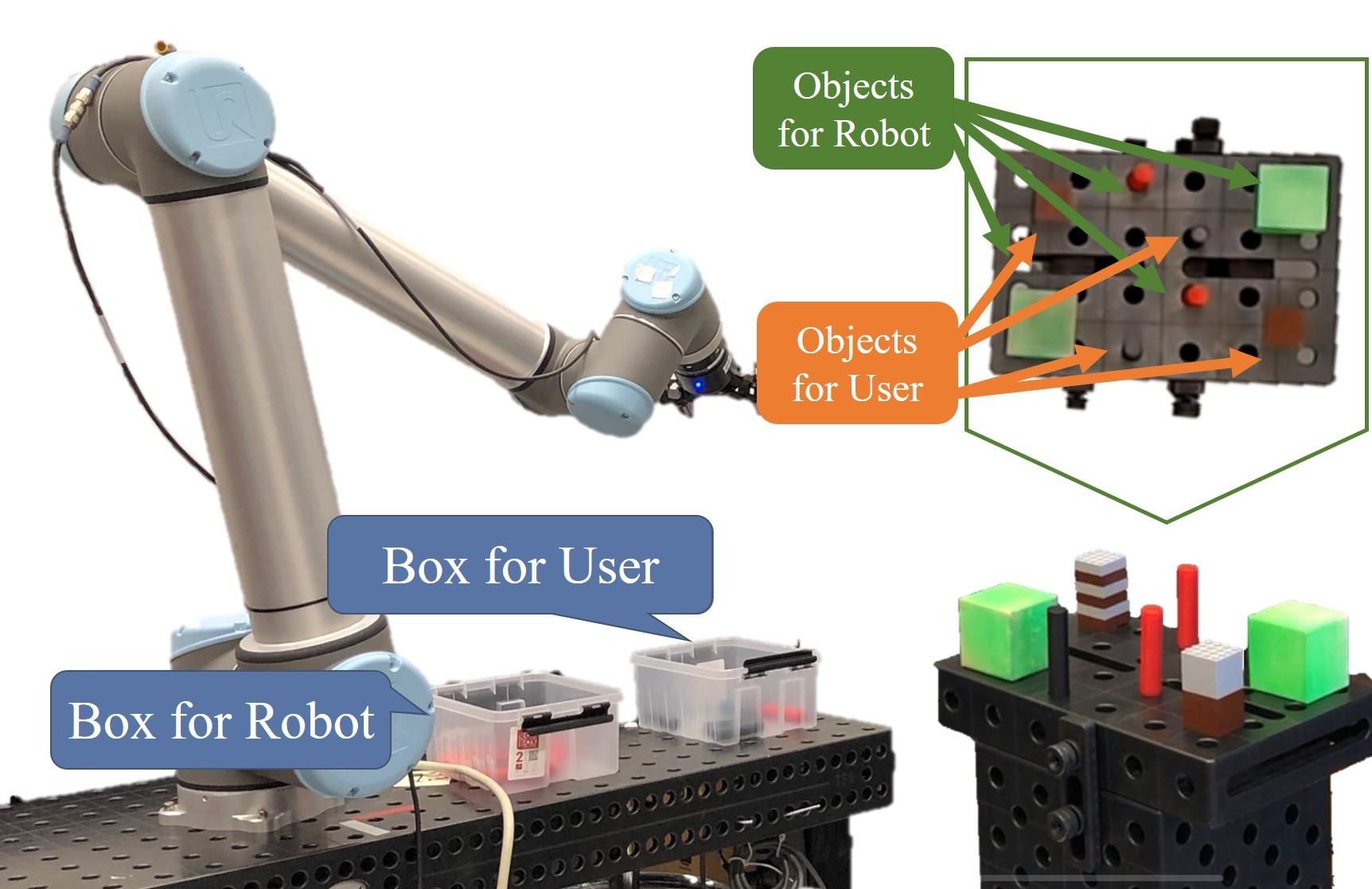}
  \qquad
  \vspace*{-5mm}
  \caption{Experimental setup. Robot takes green cubes and red cylinders. The user takes brown cubes and black cylinders. Then they place cubes and cylinders in their corresponding boxes. }
  \vspace{-4mm}
  \label{fig:EnviromentExp3}
\end{figure}   

The goal of the experiment was to evaluate the system performance during human-robot collaboration in a shared working environment. This experiment resembles common activities in a warehouse. The task for the robot (6 DoF collaborative robot UR10 with a 2 fingers gripper from Robotiq) was to put objects (two cubes and two cylinders) from a rack into boxes. The path of the robot TCP was recorded one time without human-robot interaction. The user's task was to assemble two cubes of plastic construction toys Lego in different color patterns and to measure the length of two cylinders, with the goal to spend more time in the same shared workspace as the robot. The user was located in front of the collaborative robot UR10 and was asked to wear, in the first trial, the Antilatency 6 DOF camera on the right hand without haptic feedback, and, in the second trial, to wear the haptic device $RecyGlide$ on the right forearm to activate the haptic feedback.

The critical distance of activation $d_{ACT}$ was set at 10 cm, and the avoidance threshold distance $d_{AT}$ was set at 30 cm. The maximum speed of the robot was set at 0.2 m/s, and the data collection frequency was set to 10 Hz. The haptic display was set to render the distance from the robot to the user hand $d_{RO}$ in a range of 40 cm to 20 cm to avoid being closer than the critical distance of activation, and to analyze the behavior of the avoidance algorithm during the movement of the user hand.  

\begin{figure}[h!]

  \centering
  \includegraphics[width=0.47\textwidth]{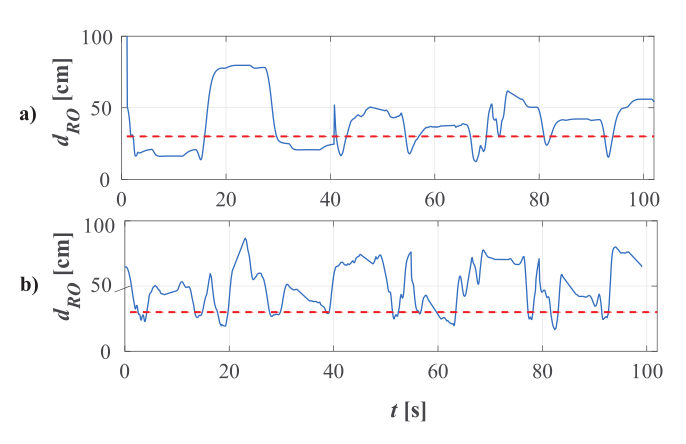}
  \qquad
  \vspace*{-0.8cm}
  \caption{Human-Robot Interaction Experiment in a Collaborative Task. Distance between the robot's TCP and the user's hand without haptic feedback (a), and with haptic feedback (b). The red line represents the avoidance threshold distance $d_{AT}$. }
  \label{fig:ex3}
  \vspace{-3mm}
\end{figure}

\begin{figure}[h!]
  \centering
  \includegraphics[width=0.47\textwidth]{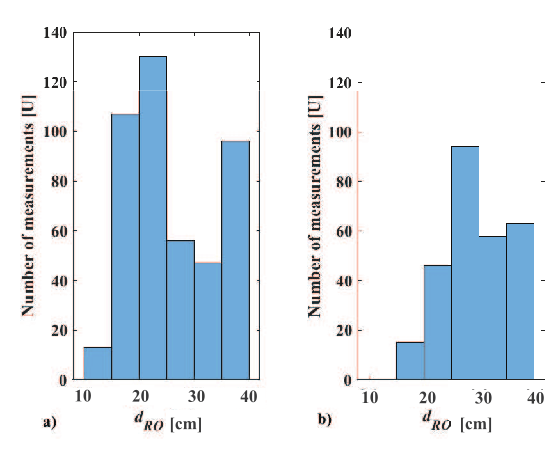}
  \qquad
  \vspace*{-5mm}
  \caption{Measurements of robot-obstacle distance inside of the range from 0 to 40 cm for the collaborative task without haptic feedback (a) and with haptic feedback (b).}
  \label{fig:dmin5}
\end{figure}

\textbf{Results}: Fig. \ref{fig:ex3} shows the robot-obstacle distance $d_{RO}$ without haptic feedback (a) and with haptic feedback (b). The minimum robot-obstacle distances are 12.4 cm (without haptic feedback) and 16.6 cm (with haptic feedback). The red line in Fig. \ref{fig:ex3} represents the avoidance threshold distance $d_{AT}$, which means that every time that the distance is under the red line in Fig. \ref{fig:ex3}(b), the haptic feedback was active. Fig. \ref{fig:dmin5} shows the histogram of the robot-obstacle distance samples for distances less than 40 cm, which corresponds to the distance when the haptic feedback starts to render information. We can observe that the number of measurements with haptic feedback is less than those without haptic feedback. The peaks of the number of measurements without haptic feedback are between 20 and 25 cm, while those with haptic feedback are located between 25 - 30 cm, which means that the haptic feedback helped the user to stay farther from the robot TCP. The 38\% of points are under 30 cm ($d_{AT}$) without haptic feedback and 19\% under 30 cm with haptic feedback. 

Using $RecyGlide$, the robot-obstacle distance $d_{RO}$ increases, and the time that the user is inside of the avoidance area decreases by 50\%. These results show that, using the haptic device, the user can avoid the robot TCP, reducing the security hazard in an industrial environment, and decreasing the time of the operation.

\begin{figure}[h!]
  \centering
  \vspace*{0.2cm}
  \includegraphics[width=0.52\textwidth]{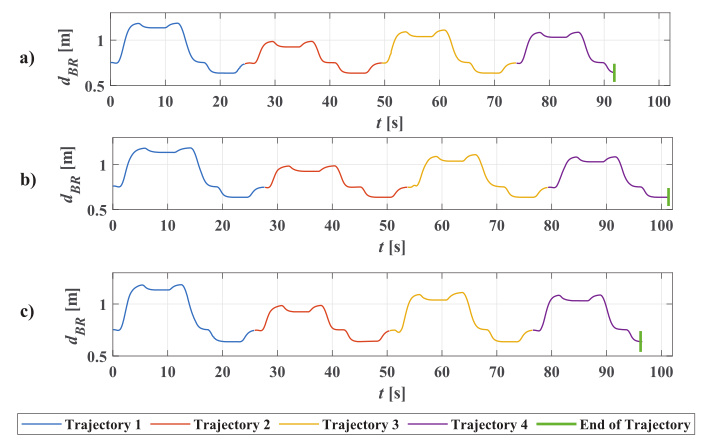}
  \qquad
  \vspace*{-5mm}
  \caption{Human-Robot Interaction Experiment in a Collaborative Task. The diagram represents the TCP trajectories without obstacle (top), with obstacles without haptic feedback (middle), and with obstacles and haptic feedback (down) vs. time. The trajectories 1, 2, 3, and 4 represent TCP path to pick up each target (cube 1, cube 2, cylinder 1, and cylinder 2) and to transport them to the collecting box.}
  \label{fig:u}
\end{figure}

Fig. \ref{fig:u} shows the distance from the TCP to the robot base. The employed time for the collaborative task without haptic display (Fig. \ref{fig:u} b)) is higher. The times employed for collision avoidance are 9.44 s (without haptic feedback) and 4.35 s (with haptic feddback). When the user wears the haptic display, the time when the collision avoidance algorithm was active reduced by 50\%.

The recorded paths by the robot are 19.37 m (without human-robot interaction), 19.75 m (human-robot interaction without haptic feedback), and 19.44 m (human-robot interaction with haptic feedback). The collision paths (additional recorded path by the robot to avoid obstacles) are 38 cm (without haptic feedback)  and 7 cm (with haptic feedback). Overall, these results indicate that, with haptic feedback,  the robot's path when the collision avoidance algorithm is activated was reduced by 81.49\%. 

\section{Conclusions and Future Work}

This research proposes a novel Human-Robot collaboration framework which features a wearable multi-modal haptic display $RecyGlide$, wearable motion capturing system, and intelligent collision avoidance algorithm.

The wearable haptic display aims to inform the user about her/his distance to the robot TCP and the robot collision avoidance algorithm mode. Two experiments were performed to evaluate human behavior to the stimuli, and the collision avoidance algorithm in an imminent collision with and without haptic feedback. A third experiment evaluated the system during a collaborative task. 

The first experiment revealed that a blind user is capable of avoiding a static point using only haptic feedback. According to the last experiment, the haptic stimuli help the user to distance itself from the robot TCP, reducing the activation of the robot collision avoidance algorithm, decreasing the operation time, and improving human safety.

This study has shown that when haptic feedback in the control loop was included, the robot-obstacle minimum distance increased by 4.1 cm, and the robot's path, when the collision avoidance algorithm was activated, was reduced by 81\%.  The proposed framework performs a redundant avoidance because the human and the robot repeal each other. The user is more aware of what is happening in the environment and avoids the robot's position. 

In future work, the proposed algorithm will be tested for an assembly task. It will be important to achieve whole body-robot safe interaction. Therefore, the research will be also devoted to the development of wearable tracking system to calculate the coordinates of the feature points of human body skeleton. 

The proposed technology CoHaptics will be merging the strength of human and robot: the robot has high payload and high repeatability while the human operator possesses versatility and manual skills in dexterous activities. The proposed algorithm will enable a safe human-robot collaboration with a low risk of collision. 

\section*{Acknowledgement}
The reported study was funded by RFBR and CNRS, project number 21-58-15006 and the project ``Energy-efficient Programming of Collaborative Robots” funded by ELFORSK.

\addtolength{\textheight}{-12cm}   






\end{document}